\documentclass[runningheads]{llncs}
\usepackage{graphicx}
\usepackage{xcolor}
\usepackage{hyperref}

\begin{document}
%
\title{The Landscape of Ontology Reuse Approaches}
%
%
\author{Valentina Anita Carriero\inst{1} \and
Marilena Daquino\inst{1} \and
Aldo Gangemi\inst{1,2} \and \\
Andrea Giovanni Nuzzolese\inst{2}\and
Silvio Peroni\inst{1}\and
Valentina Presutti\inst{1,2}\and
Francesca Tomasi\inst{1}\thanks{Authors’ contributions: Gangemi A., Presutti V. and Tomasi F. (section 1); Carriero V. (section 2 and 4.2). Daquino M. (section 3.2 and 3.4), Nuzzolese A. (section 3.1 and 3.3). Peroni S. (section 4.1). All authors contributed to section 5.}}

\authorrunning{V.A. Carriero et al.}

\institute{University of Bologna, Italy
\email{\{valentina.carriero3,marilena.daquino2,aldo.gangemi, silvio.peroni,valentina.presutti,francesca.tomasi\}@unibo.it} \and
Institute of Cognitive Sciences and Technologies, National Research Council, Italy\\
\email{\{andrea.nuzzolese\}@cnr.it}}
\maketitle             
\begin{abstract}
Ontology reuse aims to foster interoperability and facilitate knowledge reuse. Several approaches are typically evaluated by ontology engineers when bootstrapping a new project.
However, current practices are often motivated by subjective, case-by-case decisions, which hamper the definition of a recommended behaviour. In this chapter we argue that to date there are no effective solutions for supporting developers' decision-making process when deciding on an ontology reuse strategy. The objective is twofold: (i) to survey current approaches to ontology reuse, presenting motivations, strategies, benefits and limits, and (ii) to analyse two representative approaches and discuss their merits. 

\end{abstract}

\section{Introduction}
%
%
%
Ontology Reuse (OR) is a critical aspect for the evolution of the Semantic Web since its origins \cite{gruber1993translation,neches1991enabling,studer1998knowledge,katsumi2016ontology}. OR aims to foster semantic interoperability between data sources and facilitate data integration tasks. As a consequence, it is naturally intertwined with other ontology engineering areas, such as ontology selection (OS), ontology integration (OI), and ontology access (OA), which in turn affect OR strategies.

To date, several approaches to OR exist, due to different motivations or different implementations. For instance, institutions and community consortia typically encourage \emph{direct} reuse of existing standard or popular ontologies. Research communities and communities of practice also suggest \emph{indirect} approaches, i.e. by designing ontologies tailored to a use case and aligning them to standard or popular ones when appropriate, or \emph{hybrid} approaches, i.e. when mixing direct and indirect approaches. 

The decision on the OR approach to be adopted is usually done by ontology engineers when bootstrapping a project. However, despite a vast literature on OR provides definitions and requirements, decision methods may be biased and shallow. Hence, solutions are often taken on a case-by-case basis, hampering the definition of shareable good practices. 
On the one hand, reuse approaches are primarily accompanied or motivated by design methods. Those are ultimately based on the conceptualisation needs of an ontology project. Methods borrowed from software engineering, philosophy and cognitive science have emerged. Examples are competency questions \cite{Gruninger1994}, foundational ontologies \cite{gangemi2002sweetening}, task-oriented quality assessment \cite{DBLP:conf/esws/GangemiCCL06}, design patterns \cite{Gangemi05,Gangemi2009,DBLP:books/ios/HGJKP2016}, etc. Such methods aim at making explicit the general cognitive requirements that should guide an ontology project. 
On the other hand, social argumentations often affect designers' decisions. International institutions, community consortia and standardisation bodies tend to support their own practices and shared ontologies; domain experts may perceive the reuse of known ontologies as simpler and more authoritative; ontologists tend to prioritise their expertise, etc.



In this chapter we argue that to date only a few  solutions exist (e.g. \cite{hyland2014best,Presutti2016role}) for making pragmatic decisions about which OR approach should be adopted in the development of a new ontology, and a more comprehensive effort is needed. 
The aim of this work is twofold: (i) to provide a comprehensive account of existing solutions, addressing benefits and limits, and (ii) to support developers’ decision-making process by discussing two representative real-world scenarios in the light of general considerations addressed in (i).  

The chapter is organized as follows. In Section \ref{approaches} we describe motivations underlying OR approaches and strategies of OI. In Section \ref{challenges} we present the state of the art, benefits, and limits of current practices in ontology selection OS, ontology access OA, ontology integration OI, and OR implementation. In Section \ref{casestudies} we present two case studies adopting different OR approaches. In Section \ref{discussion} we discuss choices made in such contexts in the light of their portability. We conclude with a summary of dimensions to be taken into account when evaluating OR approaches.

\section{Approaches to ontology reuse} \label{approaches}

When bootstrapping a new project, developers search and select candidate ontologies to be reused, and decide strategies for OI. In this section we introduce common motivations guiding OS, policies for OR, and methods for OI. 

\paragraph{\textbf{A. Motivations guiding ontology selection}.}
Some guidelines for Linked Open Data and vocabularies publication \cite{hyland2014best,berrueta2008best} support developers in the selection of valid and documented ontologies, by promoting either \emph{top-down agreement} as implemented by standards, or \emph{currency}, by reusing  popular ontologies. The designer of an ontology decides that its requirements are fully (or mostly) identical to those that have inspired a standard or a popular ontology.

Research communities  \cite{Gruninger1994,gangemi2002sweetening,DBLP:books/daglib/0028799,DBLP:books/ios/HGJKP2016,Presutti2016role} and some communities of practice as with some W3 Consortium recommendations (\url{https://www.w3.org/2001/sw/BestPractices/OEP/}) foster \emph{explicit cognitive analysis} in OR, notably the indirect reuse of ontology patterns rather than ontology terms.

\textbf{OR by standardisation.} OR by standardisation refers to the practice of reusing ontologies issued by authoritative organisations, like ISO, W3 Consortium, and professional or community consortia. Examples include ISO standard ontologies such as CIDOC-CRM \cite{cidoc2017}, W3C endorsed ontologies such as PROV-O \cite{lebo2013prov} or Time Ontology (\url{https://www.w3.org/TR/owl-time/}), and Community standard ontologies such as FRBRoo \cite{riva2017frbroo}. Standard ontologies provide reference models for representing cross-domain information, e.g. events, temporal aspects, provenance, and often provide domain-oriented models meant to allow stakeholders to perform lossless data transformation into RDF. 
Standard ontologies are usually recommended for direct reuse to members of a community, or can be intended as reference documents, envisioning specialisations and extensions.
  
\textbf{OR by popularity.} OR by popularity refers to the practice of reusing ontologies that are popular, typically because they are reused in (i) many third-party ontologies \cite{hyland2014best} by import, specialisation or extension, and in (ii) existing datasets by instantiation of ontology terms \cite{sabou2006ontology}. Popular ontologies might be designed by standardisation bodies, or might include cognitive analysis, but that is probably not the main reason they are reused. For example, FOAF (\url{http://xmlns.com/foaf/spec/}) is the result of an early exemplification (1999) of a vocabulary for the Semantic Web, which did not emerge (did not intend to do so) from any standardisation or cognitive/formal analysis. Nonetheless, it has become a reference for semantic social network data. Another example is Good Relations \cite{hepp2008goodrelations}, which has been designed to address an important interoperability problem in eCommerce. It is not a standardisation result, but is based on formal requirements from industry and cognitive analysis. Other popular ontologies include DBpedia (\url{http://dbpedia.org/ontology/}), the result of collective cleaning of Wikipedia user-based template specifications, and schema.org (\url{https://schema.org/}), an output of an industrial standardisation body fostering search engine optimisation via semantics.

\textbf{OR by cognitive analysis.} OR by cognitive analysis refers to the definition of requirements of an ontology project as primary source of decisions. that may lead to novel constructs, or the usage of existing design components, such as foundational ontologies \cite{gangemi2002sweetening} or ontology patterns \cite{DBLP:books/ios/HGJKP2016}. It is associated with measures to justify the axioms contained in an ontology and supports decisions about the reuse or the novel design of axioms. In particular, whether to reuse something or not is supposed to be dependent from the requirements of an ontology project, both functional and non-functional (e.g. usability, multilingualism). 
As aforementioned, cognitive analysis borrows methods from software engineering, philosophy and cognitive science, the most popular including {\em competency questions} (CQs) \cite{Gruninger1994}, formal ontological analysis using patterns already encoded in foundational ontologies, e.g. DOLCE \cite{gangemi2002sweetening}, task-oriented quality assessment \cite{DBLP:conf/esws/GangemiCCL06}, etc. A community has emerged to integrate and support cognitive analysis by means of Ontology Design Patterns (ODP, \url{http://www.ontologydesignpatterns.org}) \cite{Gangemi05,Gangemi2009,DBLP:books/ios/HGJKP2016}. According to~\cite{Gangemi2009}, an ODP is a modelling solution to solve a recurrent ontology design problem. ODPs show certain characteristics, i.e. they are: computational, small, autonomous, hierarchical, cognitively relevant, linguistically relevant, and best practices. Furthermore, an ODP is intuitive and compact, and catches ``core'' notions of a domain. Notions are gathered in sets of CQs, which are the key tool for designing and enabling reuse of ODPs.

\paragraph{\textbf{B. Policies and implementation strategies}.}
Policies for implementing OR take into account motivations and requirements of the project, developers' design choices, and constraints given by the state of the art technologies. We distinguish policies for OR as follows: direct reuse, indirect reuse (as in \cite{Presutti2016role}), and hybrid reuse.
\emph{Direct reuse} may be performed in two ways, namely: (i) the import of ontologies into a new ontology (i.e. by means of the axiom \texttt{owl:imports}); (ii) the inclusion of selected ontology terms in a new ontology (possibly referencing reused ontologies by means of e.g. \texttt{rdfs:isDefinedBy} axioms). In the former case, the semantics of reused ontologies is included into the new one. In the latter, the semantics of reused terms is delegated to external ontologies.
\emph{Indirect reuse} implies that terms from external ontologies are reused as templates, or just aligned, in the new ontology. Terms and their semantics (axioms) are natively described in the new ontology, and aligned to reused terms by means of e.g. \texttt{rdfs:subClassOf} or \texttt{owl:equivalentProperty} axioms.
\emph{Hybrid reuse} is a design choice where ontology terms are selected either for direct reuse or for being indirectly reused as templates, according to characteristics of reused ontologies and requirements of the project. 

Actual methods for integrating ontologies depend on the extent of reusable knowledge and the extent of required changes.
When requirements to be addressed by the new ontology are largely satisfied by existing ontologies, the latter can be reused \emph{as-is}. However, some manipulation of existing ontologies in the new setting is likely to be required \cite{Simperl2009}. Possible changes - e.g. extension, adaptation, specialization, change of naming conventions - may be needed due to the heterogeneity or incompleteness of reused ontologies, or the lack of sufficient abstraction or specificity. Moreover, ontology harmonization tasks or orthogonality of terms may be required.

Two methods are applicable, namely: composition (also referred to as integration), and merging (also known as fusion) \cite{pinto2004ontologieshow}. 
\emph{Modular composition} is the practice of reusing ontology modules, i.e. sets of terms and axioms that address a specific subset of requirements, into the new ontology. Reused ontologies usually cover different subjects, and are likely to be orthogonal, thus there is little overlap of concepts and semantics. A particular case of composition is when an ontology is built by integrating Ontology Design Patterns (ODPs). In this case, the fragment tackling specific requirements is clearly and formally defined in a dedicated ontology, and has been explicitly designed for reuse.
\emph{Merging} refers to a process where individual concepts, axioms, and statements from reused ontologies are merged together into a new model. In this case, there are several overlapping areas, since the ontologies are on the same subject, and harmonization tasks are required.
While modular composition allows to clearly identify regions or modules reused from source ontologies, merging terms does not allow to immediately understand which source contributes to address a representational issue (e.g. a competency question), as individual concepts may loose or change their semantics in the context of the new ontology.

\section{Ontology reuse: benefits, gaps, and challenges}\label{challenges}



In this section we survey the state of the art of current OR practices and we address benefits and challenges  affecting OR decisions. We loosely adopt FAIR principles \cite{wilkinson2016fair} as a framework to characterise aspects that affect available ontologies and OR tasks, namely: (i) selection (relevant to findability of ontologies), (ii) access and preservation (relevant to accessibility), (iii) integration (relevant to interoperability) and (iv) implementation (relevant to reusability). 

\subsection{Ontology selection}

\textbf{State of the art.} The number of existing ontologies increased significantly over the last years, and tools to support OS have become a fundamental aid. Existing tools include: (a) catalogues of general purpose ontologies, such as Linked Open Vocabularies (LOV) \cite{vandenbussche2017linked}, vocab.org (\url{http://purl.org/vocab/}), ontologi.es (\url{http://ontologi.es/}), and Joinup platform (\url{https://joinup.ec.europa.eu}); (b) catalogues of domain ontologies, e.g. BioPortal (biomedicine) \cite{whetzel2011bioportal}, SOCoP+OOR (geography, \url{https://ontohub.org/socop}), and SWEET (Earth and environment) \cite{raskin2005knowledge}; (c) catalogues of ODPs \cite{presutti2008content}; (d) semantic web search engines, e.g. Watson \cite{d2007watson}, Falcons \cite{cheng2008falcons}, prefix.cc (\url{http://prefix.cc/}), Swoogle \cite{finin2005swoogle}, and Schemapedia (\url{http://schemapedia.com/}). 
Services support users in browsing, filtering and searching over ontologies metadata and terms \cite{dAquin12}. Swoogle and LOV also show ranking scores of ontology terms, based on their popularity across LOD datasets. Moreover, LOV offers users' reviews and insights on the relatedness of ontologies, e.g. imports, extensions or specialisations, to support ontology alignment.

\textbf{Benefits.} Popularity-based metrics are common when searching for ontologies and can avoid time-consuming selection activities when looking for general purpose ontologies. Moreover, reusing popular ontologies increases the chances that data will be reused by other applications~\cite{Heath2011}. Such a practice is very common in the Linked Data community. It is essentially driven by the assumption that the semantics of an ontology is primarily based on the intuitiveness of their names and their popularity in the Web of Data.
Reusing well-known ontologies fosters semantic interoperability and homogenization at Web scale. In fact, re-using well established ontological solutions allows the construction of a shared understanding over the Web knowledge with a decentralised approach. 
Such a scenario facilitates rational agents to leverage the Web knowledge and consequently interact at the knowledge level as envisioned by~\cite{Newell1982}. Indeed, building rational agents able to automatically share, reuse, exchange, and reason on the knowledge represented at Web scale is the El Dorado of the Semantic Web Community~\cite{Berners-Lee2001}.

\textbf{Gaps and challenges.} To date, OS is a highly subjective task, often manually performed by experienced ontology engineers, who select ontologies that intuitively fit for the purpose. 
Secondly, only popularity-based metrics are available, which may foster biased behaviours in OR practices, such as boosting OR by popularity and OR by standardisation only. A correct OS must rely on clear ontological requirements gathered by analysing the modelling problem and the domain. In literature~\cite{Gruninger1994} ontological requirements are commonly identified by competency questions. The latter are the key tools that enable many ontology design methodologies based on ontology reuse. A notable example is eXtreme Design (XD)~\cite{Blomqvist2016}.
Moreover, internationalisation \cite{Gracia12} and licensing for reuse \cite{Poblet16} should be taken into account, as well as more sophisticated statistics on the co-occurrence of ontologies in third-party ontologies and datasets. The latter would show how ontologies are combined (i.e. by composition or merging), shedding light on orthogonality and harmonisation between ontologies. Detailed information on the extent ontologies are actually reused in other ontologies (e.g. whether single classes, properties or modules) along with information on the application domain, would help to classify reusing ontologies in families of ontologies, assuming that ontologies in a certain domain share commonalities \cite{Ochs17}.
Lastly, clustering datasets where ontologies co-occur would allow to classify ontologies on the basis of their application domain. While statistics on the LOD cloud are available \cite{Ermilov13}, and instance-level relations in data can be discovered \cite{Asprino19}, more sophisticated systems for ontology usage tracking are not available, and consequently supporting ontology selection is still a manual, error-prone activity.

\subsection{Ontology access and preservation}

\textbf{State of the art.} Third parties that reuse ontologies create a dependency on the original vocabulary that includes the semantics of reused terms. Hence, access, maintenance, and long-term preservation of ontologies are critical aspects for the development of new ontologies. Moreover, access methods must address versioning, since ontologies may evolve over time.

Well-known services to ensure persistence of HTTP URIs and correct content negotiation - e.g. purl.org (\url{https://archive.org/services/purl/}), w3id.org (\url{https://w3id.org/}) - and methods to access ontology contents - e.g. dump an OWL file, query a SPARQL endpoint, send requests to APIs - are available. Popular vocabularies such as schema.org or FOAF are currently maintained on version-controlled repositories, e.g. GitHub. Several efforts were made to support ontology versioning in the ontology development phases \cite{garijo2017widoco,halilaj2016vocol}.

In 2012 Poveda-Villalón et al.
\cite{poveda2012landscape} semi-automatically accessed and analyzed 256 ontologies included in LOV registry, showing that 23\% of surveyed ontologies were not available. In 2018 Fèrnandez-Lopez et al. \cite{fernandez2019ontologies} performed a similar investigation, showing that the percentage increased to 36\%. 

\textbf{Benefits.} Relying on reused ontologies delegates to third-party ontology engineers the problem of dealing with ontology preservation, versioning, storage and evolution. If an ontology engineer opts for the reuse of ontologies that result from well established initiatives (e.g. FOAF, Schema.org) they might count on large communities to cope with both maintenance and preservation, along with the compliance with standards (e.g. serialisation syntaxes, modelling languages, etc). The latter are part, as the ontologies themselves, of the evolutionary process that affects any human artifact. For example, new serialisation formats may evolve or arise. Similarly, new logical patterns might be included in the reference modelling language (e.g. punning has been introduced in OWL 2, but it was not available in OWL 1).

\textbf{Gaps and challenges.} Aforementioned surveys show that there is an urge for vocabulary publishers to ensure long-term availability of ontologies and allow programmatic processing by catalogues. However, there is no standard solution to ensure long-term preservation of vocabularies on the Web \cite{baker2013requirements}. Free of charge proprietary repositories are currently the cheapest solution for sharing source code, but they lack some features that systems dedicated to distributed vocabulary development usually provide, e.g. integrated support for modeling, population and testing \cite{halilaj2016vocol}. Moreover, maintenance and availability are ensured by suppliers only, arising trust issues. As a matter of fact, ontologies produced by small enterprises and scholars in the context of short-term research projects are likely to suffer long-term preservation issues. 

\subsection{Ontology integration}

\textbf{State of the art.} OI methods ensure different degrees of interoperability between reused and reusing ontologies depending on policies, design choices, and whether the process is manually or automatically performed.

Several frameworks for semi-automatic ontology aggregations have been proposed, such as architectures for automating ontology generation through ontology reuse \cite{lonsdale2010reusing}, ranking models for evaluating ontologies based on semantic matching \cite{park2011ontology}, as well as compatibility metrics for comparing and integrating ontologies \cite{trokanas2016ontology}, frameworks for ontology integration based on knowledge graphs integration and machine learning techniques to identify core terms to be integrated \cite{zhao2014ontology}, and methods for building ontologies from ontology patterns and vice-versa \cite{ruy2017reference}. Ontology integration based on ontology matching has been largely investigated in literature~\cite{Euzenat2011,Euzenat2013} and many solutions exist at the state of the art, such as~\cite{Jain2010,David2011}. 

\textbf{Benefits.} Ontology matching relieves ontology engineers of the cognitive effort of identifying alignments manually and speeds up the ontology modelling process. Additionally, it allows ontology engineers who are not domain experts to design ontologies by dealing with fine-grained terminologies wrt the domain. 

The ontology design process might benefit of multiple integrations. For example, most of the solutions at the state of the art associate more than one correspondence with external ontologies for each class or property the ontology engineer aims at integrating within the target ontology.
Finally, the requirement of reusing well established and largely adopted ontologies is transparently addressed by following this approach. In fact, ontology matching algorithms might be fed with repositories and catalogues referencing specific target ontologies that address the latter requirement.


\textbf{Gaps and challenges.} Integration activities may be error-prone. Ambiguities, inconsistencies, and heterogeneity in existing ontologies may affect results at different dimensions, such as syntactic heterogeneity, terminological heterogeneity, conceptual heterogeneity, and semiotic heterogeneity \cite{Euzenat2013}. Gangemi and Pisanelli \cite{gangemi1998ontology} previously described terminological issues of local ontologies in terms of semantic imprecision (e.g. relation range violation), ontological opaqueness (e.g. lack of motivation for choosing a certain predicate), and awkward linguistic naming conventions. Hence, human intervention is likely to be required not only in the search and the selection of ontologies, but also in the disambiguation and formalisation of integrated ontologies. 

Secondly, ontology integration is not a one-time task. It may need to be applied several times, since integrated ontologies may change over time \cite{pinto2004ontologieshow}. According to the literature, it is not clear yet whether it is more cost-effective to build a new ontology from scratch that perfectly meets current needs than to try to rebuild and adapt existing ontologies.

\subsection{Ontology reuse implementation}

\textbf{State of the art.} Ontologies resulting from OR practice may reference and acknowledge original ontologies in several ways \cite{poveda2012landscape}, such as (1) importing the entire ontology, (2) reference ontology terms by means of annotation properties, or (3) acknowledging reused ontology in external alignment documents.

In the empirical survey conducted by Schaible et al.
\cite{schaible2014survey} on 79 LOD practitioners, the authors show that direct reuse of popular vocabularies is preferred than defining new terms and align to other vocabulary terms. In \cite{fernandez2019ontologies} the authors compare results of their analysis on OR trends in LOV ontologies to the previous work done by \cite{poveda2012landscape}. They show that W3C endorsed ontologies are reused with different nuances in 78\% of ontologies catalogued in LOV catalogue. More important, they notice that the evolution in OR practices from direct import of ontologies to the direct reuse of ontology terms has increased until approximately a half of the total re-users. OR practices in domain applications confirm the same trend. Experiments in the biomedical domain showed that OR between ontologies from BioPortal is limited, and presents a number of antipatterns \cite{kamdar2017systematic}. \cite{Ochs17} shows that the same ontologies available in BioPortal adopt several nuances of direct reuse, including specializations and extensions.

\textbf{Benefits.} On the one hand, \texttt{owl:imports} axiom is natively supported by OWL and its semantics is clear. Additionally, to the best of our knowledge, most ontology development frameworks (e.g. Prot\'{e}g\'{e}) support the axiom.
On the other hand, the use of annotation properties, e.g. \texttt{rdfs:isDefinedBy}, introduces more flexibility with respect to modularity (i.e. it is possible to declare which parts of an external ontology are reused rather than importing the whole ontology).  
More recently, a few solutions have been proposed for implementing OR by relaxing the monolithic constraint of \texttt{owl:imports} and preserving formal semantics at the same time. For example, OPLa~\cite{Hitzler17opla} defines classes and properties for declaring which modules or patterns are reused within a target ontology. OPLa is supported by Prot\'{e}g\'{e} by means of a dedicated plug-in~\cite{Shimizu2018}.

\textbf{Gaps and challenges.} On the one hand, \texttt{owl:imports} does not allow to customise the import of ontologies (i.e. it only allows to import ontologies as a whole).
As reported in \cite{hammar2012}, being \texttt{owl:imports} transitive, concepts included in unintentionally imported ontologies may be irrelevant to and incompatible with the requirements of the local ontology.
Moreover, when importing ontologies into a new one, it may happen that the reused ontologies are not available over time, and their semantics is lost.
On the other hand, the flexibility given by annotation properties corresponds to less semantic rigour. As a matter of fact, none of the built-in annotation properties is explicitly meant for ontology reuse. That is, their adoption for implementing reuse does not address any formal semantics. Hence, none of the existing platforms for ontology design fully support them. In fact, when only terms belonging to external ontologies are imported in a new ontology, reasoning on the reused ontology constraints is not possible when reasoning on the reusing one. Secondly, if a reusing ontology includes contradictions with respect to a reused ontology, these cannot be automatically detected. Only when both ontologies are imported in a third ontology, contradictions may emerge \cite{fernandez2019ontologies}. Likewise, when external documents including alignments are provided, the same situation may happen if adopters do not include alignments in the reusing ontology.

\section{Case studies}\label{casestudies}

In this section we introduce two representative use cases exemplifying respectively direct reuse and hybrid reuse approaches. 

\subsection{Direct reuse of ontologies: the OpenCitations Data Model}

\begin{figure}
\includegraphics[width=\textwidth]{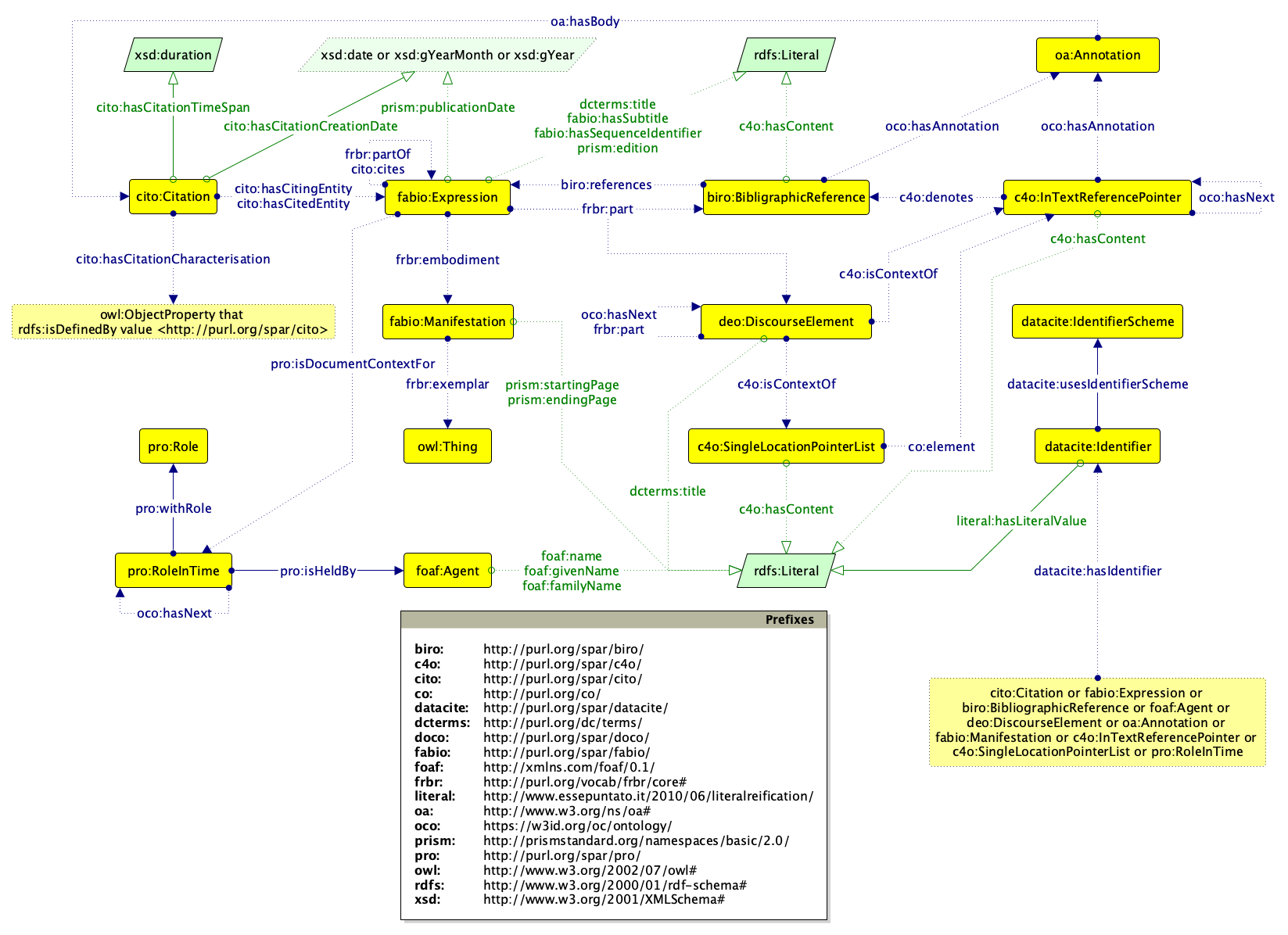}
\caption{Overview of OCDM main classes and properties} \label{oco-fig1}
\end{figure}

The OpenCitations Data Model (OCDM) \cite{peroni2018opencitations} is based on the OpenCitations Ontology (OCO, \url{https://w3id.org/oc/ontology}), which aggregates terms from existing well-known ontologies. It was initially developed in 2016 by the OpenCitations organisation to describe data in the OpenCitations Corpus (\url{http://opencitations.net/corpus}). In recent years OpenCitations has developed other datasets, and the OCDM is currently adopted by several external projects that contribute to the growth of the model. The current version of OCDM takes such changes into account and describes a generic bibliographic dataset, making easier its adoption by third parties. Fig. \ref{oco-fig1} shows classes and properties of OCO.

\textbf{Ontology development methodology.}
The first version of the OCDM released in 2016 was developed by directly reusing (i.e. merging) terms from existing models. The latter include the SPAR Ontologies, a set of modular and orthogonal ontologies developed by using SAMOD \cite{peroni2016simplified}, an agile data-driven methodology for ontology development.
Within the context of a project recently funded by the Wellcome Trust, SAMOD was adopted to extend OCO with terms relevant to generic bibliographic datasets. The outcome includes motivating scenarios, competency questions, and a glossary of terms of all the new entities included in OCO. 

\textbf{Ontology selection.} OCO reuses selected terms from popular domain ontologies, i.e. the SPAR Ontologies \cite{peroni2018spar}, and W3 endorsed ontologies, such as the Web Annotation Ontology (OA) \cite{sanderson2013open}, PROV-O \cite{lebo2013prov}, and VoID \cite{alexander2011describing}. The choice is motivated by several pragmatic factors, such as: ontology designers' background knowledge with respect to reused ontologies, the fact that potential stakeholders of OpenCitations datasets already use the same ontologies, and, in most of cases, the possibility to directly modify reused ontologies rather than creating new terms.

\textbf{Ontology access and preservation.} OCDM is released under CC-BY license. Both OCO and reused SPAR ontologies are developed and maintained by the OpenCitations organisation. Long term preservation of such artefacts is ensured by the OpenCitations organisation, that has been recently selected by the Global Sustainability Coalition for Open Science Services (SCOSS, \url{https://scoss.org/}) for its funding programme that guarantees long term sustainability. Accessibility of other reused ontologies is ensured by W3 Consortium.

\textbf{Ontology integration.} The integration process of ontological terms gathered from aforementioned ontologies led to the creation of a new single ontology (i.e. OCO) with the purpose of aggregating only the terms that are relevant to the representation of motivating scenarios. Few extensions and specialisations of reused ontologies were necessary. Whenever applicable, changes were directly applied in the reused ontologies, e.g. new terms were added to SPAR Ontologies. Only in one case an initial request\footnote{See \url{https://lists.w3.org/Archives/Public/public-openannotation/2019Sep/}} for the addition of a new property to a W3 endorsed ontology (i.e. OA) was made. Meanwhile, a bespoke property \texttt{oco:hasAnnotation} has been created and declared as the inverse property of \texttt{oa:hasTarget}. 

\textbf{Ontology reuse implementation.}
As aforementioned, OCO directly reuses existing ontology terms rather than importing ontologies. Provenance of reused terms is recorded by means of \texttt{rdfs:isDefinedBy} statements.

\subsection{Hybrid reuse of ontologies and patterns: the ArCo ontologies}

ArCo (Architecture of Knowledge, \url{https://w3id.org/arco}) is a project dedicated to the publication of the open knowledge graph (KG) of the Italian cultural heritage (CH) \cite{Carriero2019}, deriving from the General Catalogue of Italian Cultural Heritage maintained by the Italian Institute of the General Catalogue and Documentation (ICCD). 
As part of the project, several ontologies for representing data about Italian cultural heritage were developed. A wide and complex domain as CH easily leads to a large ontology, and ArCo aims at modelling with a fine grain a wide range of concepts relating to 30 different types of cultural properties; this motivates the choice of a network of ontology modules, as thematically coherent sub-areas of the domain addressing a subset of requirements, rather than a monolithic design. A preliminary manual clustering of metadata in sets of topics resulted in 7 modules. The \texttt{arco} module (\url{https://w3id.org/arco/ontology/arco/}) is the root node of the network, addressing top-level distinctions and a general taxonomy, while the \texttt{core} module models general concepts, reused by the other modules (e.g. location, cultural events). Ontology modules are connected by means of import axioms.


\textbf{Ontology development methodology.} In order to cope with a huge and diverse domain, and to minimize the impact of changes in its incremental releases, ArCo adopts an agile and iterative, pattern-based and test-driven ontology development methodology, called eXtreme Design (XD) \cite{Blomqvist2016}.

The project's ontological commitment is represented by general constraints and competency questions (CQs). Requirements are provided in the form of small user stories, as scenarios and real use cases, by a growing community of customers, adopters and consumers via a Google Form, GitHub issues, mailing-list and meetups. This supports ArCo in continuously collecting new emerging requirements and extending the customer team beyond its main customer, i.e. ICCD.
These user stories are analysed and transformed into competency questions and constraints, and used for selecting suitable ODPs.
The iterative design process goes in parallel with testing and validation of ontology components against requirements, by verifying CQs coverage and model consistency.

\textbf{Ontology selection.} Modules mainly reuse ontology patterns available in the Ontology Design Patterns (ODPs) catalogue, e.g. Time-Indexed Situation, Sequence: indeed, ODPs play a central role in ArCo's design and are recommended by XD as small reusable solutions. Other ontologies have been selected because of their level of standardisation or community adoption, e.g. CIDOC-CRM, Cultural-ON, OntoPiA, or as extensions of the latter, e.g. OAEntry. 

\textbf{Ontology access and preservation.} Both ontologies and data are distributed by ICCD (\url{https://github.com/ICCD-MiBACT/ArCo}) under a CC-BY 4.0 (Attribution-ShareAlike) license. A SPARQL endpoint and additional materials (software, documentation, test suit) are available.
Ontologies directly reused by ArCo are developed and maintained by the same team (ISTC-CNR).

\textbf{Ontology integration.} The ontology integration process led to the creation of a new set of ontologies, which integrate entities, fragments and patterns from existing ontologies and ontology design patterns. 
Being ArCo an evolving project, it directly reuses only ontologies that are considered reference standards by the Italian Government, evolve rather slowly and involve ArCo's team.
When ontology terms are not directly reused, manual alignments have been performed so as to record which parts matched the project's ontological commitment.

\textbf{Ontology reuse implementation.}
Ontology terms are first designed according to the set of analysed requirements, in order to primarily formalise the semantics of the domain to be represented and address all requirements with the apt level of expressiveness. Secondly, terms from the resulting ontology are either refactored or aligned to terms of existing ontologies. 

\begin{figure}
\includegraphics[width=\textwidth]{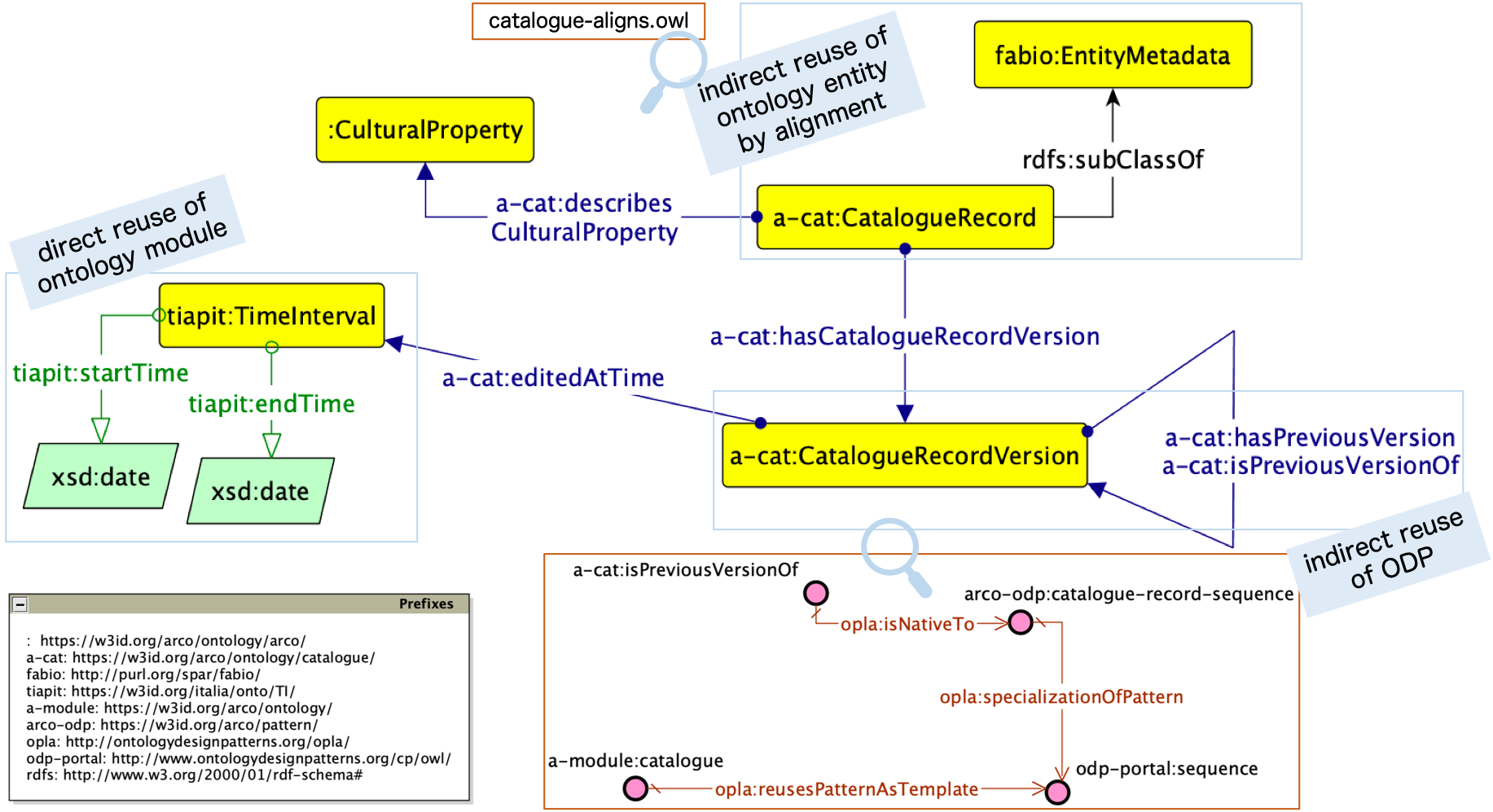}
\caption{Overview of ArCo ontology reuse approaches} \label{arco-fig1}
\end{figure}

Specifically, ArCo directly reuses Cultural-ON (\url{http://dati.beniculturali.it/cis/}), and OntoPiA (\url{https://w3id.org/italia/onto/FULL/}), a network of ontologies for top-level and cross-domain information. 
Other ontologies and Ontology Design Patterns are reused as templates. 
Alignments to external ontologies are encoded in separate files, along with documentation. Reused ODPs are annotated with the ontology design pattern annotation language OPLa (\url{http://ontologydesignpatterns.org/opla/}) 
which allows to express that new patterns are, for instance, a specialisation of existing patterns. 

An example of OR policies is shown in Figure~\ref{arco-fig1}.
In detail, terms from OntoPiA ontology, e.g. the class \texttt{tiapit:TimeInterval} and its related properties, are imported in the model. The class \texttt{a-cat:CatalogueRecord} is defined, in a separate file\footnote{\url{https://github.com/ICCD-MiBACT/ArCo/blob/master/ArCo-release/ontologie/catalogue/catalogue-aligns.owl}}, as a subclass of the class \texttt{fabio:EntityMetadata} from the FaBiO ontology.
Likewise, terms from the Sequence pattern are reused as template for new terms, e.g. \texttt{a-cat:isPreviousVersionOf}. The alignment between the new pattern and the original pattern is recorded in files annotated by means of OPLa (i.e. \texttt{opla:specializationOfPattern}). 

\section{Discussion and conclusion}\label{discussion}

In this section we discuss aspects that can support the selection of the best OR approach to be adopted at the beginning of a project. We discuss approaches adopted in the two use cases in the light of benefits and limits illustrated in Section \ref{challenges}.
Secondly, we motivate these ontology reuse approaches by using the list of attributes proposed by \cite{hammar17}, namely: implementation complexity, external reasoning dependencies, content overhead, reused content modifiability.
Finally, we draw conclusions on decision criteria and limits of this work. 

\textbf{OS.} While OCDM mainly relies on popular and W3 endorsed ontologies, ArCo mainly relies on community standards and ontology design patterns. In both cases the designers have performed cognitive analysis based on a set of competency questions.
In OCDM, existing ontologies were deemed satisfying for representing the publishing domain. Stakeholders producing similar data agree on reusing the same ontologies for the task. In ArCo, direct reuse of existing ontologies presented obstacles, such as different levels of abstraction, as in CIDOC-CRM, axiomatic paucity, as in EDM, or limits in naming conventions. In this case, stakeholders (ArCo designers, the Italian Ministry of Cultural Heritage and the larger community involved), cannot afford direct reuse.

\textbf{OA.} In both scenarios, adopters preferred to rely either on W3 endorsed standards or on ontologies developed and maintained by the same organisation that developed the new ontology. As a matter of fact, trust in third-party ontology suppliers is a fundamental aspect in OS. Secondly, both projects aim at producing and maintaining knowledge graphs that are representative of a given domain, and their long-term preservation and trustworthiness are ensured by the institutions issuing those technologies. 

\textbf{OI.} OCDM is based on OCO, a unique ontology including terms from diverse ontologies. ArCo is based on a network of ontology modules, with new terms using a new URI scheme, and appropriate alignments to existing ontologies when the semantics is the same, or more general/specific.
In the OCDM scenario, the modularity of reused ontologies is privileged, allowing adopters that are confident with well-known existing ontology definitions to retrieve faster information they seek for. Moreover, OCDM comes with a human-readable documented data model, where extensive data examples are provided as an aid. In the ArCo scenario, developers privilege the usability of new term definitions, which natively correspond to the intended conceptualisation and are under the control of the stakeholders, thus preventing any semantic shifting in external ontologies. Also in this case, there is ample documentation, Docker containers on GitHub to locally test or use ArCo data, and extensive data and query examples.

\textbf{OR.} OCDM aims at representing the publishing domain by directly reusing established ontologies, while ArCo attempts to grasp the broader cultural heritage domain that is differently represented in a vast amount of scattered projects. In the first scenario, the knowledge to be represented is stable, meaning that bibliographic data are usually factual, structured pieces of information and their representational requirements are clear. In the second scenario, ArCo is an open project: knowledge to be integrated keeps growing, and may change over time (both at instance and schema level), hence a tight control on the evolution of its semantics is needed.

\paragraph{Attributes of OR approaches.} 
Wrt the list of attributes in \cite{hammar17}, OCDM OR policy requires a low implementation effort, by directly reusing ontology terms; on the contrary, ArCo mostly adopts a more expensive approach, that is copying the structure of the reused parts into the target ontology with new IRIs. While the ArCo approach prevents from resolving external references to derive all intended inferences, also supporting content modifiability, direct reuse depends on external ontologies and generally does not allow us to modify reused content: nevertheless, this does not affect OCDM consistency and refactoring, since external reasoning is not required and the evolution of reused ontologies is under control of the OCDM team. Finally, both OCDM and ArCo have low content overhead, since only relevant content is (in)directly reused.

\paragraph{Conclusive remarks.} We reckon few dimensions in OR practices were not completely addressed in previous works. In particular, four dimensions can contribute to define a shareable set of criteria for the decision about OR approaches. The dimensions can be summarised as follows:

\begin{itemize}
    \item \textbf{Task-based OS.} Competency questions as a measure to acquire (positive or negative) requirements are currently the main intuitive, effective instrument to prevent bad design choices. This is widely accepted as a precondition for specifying conceptualisation in the form of ontologies. As such, a task-based OS guided by competency questions can justify OR approaches, followed by other common practices and motivations for OS. In addition, the community working on ontology design patterns has established good practices in matching competency questions to reusable modeling solutions, so enabling a pragmatic cognitive analysis.
    \item \textbf{Trustworthiness of OA solutions.} Political and social argumentations influence OR, such as trustworthiness of ontology suppliers that ensure OA. Such aspects apply to suppliers of both reused and reusing ontologies, and need to be pushed at the forefront of semantic technologies, as ontology design is both engineering and social negotiation.
    \item \textbf{Usability of OI results.} A trade-off between previous knowledge of ontology adopters, usability of ontology documentation, and means to facilitate query of data created according to new ontologies must be found so as to motivate OI strategies and results.
    \item \textbf{Stability of knowledge that implements OR.} Domain knowledge keeps growing and evolving, and OR practices are inevitably affected. Clear evidence of the current state of the knowledge domain should be addressed at the beginning of a project in order to justify OR implementation strategies. Here a raised awareness of cognitive analysis practices used in ontology design projects is a major priority to enhance semantic interoperability.
\end{itemize}

Decision approaches and metrics should be provided in order to effectively support ontology designers when considering the dimensions discussed. Some of those aspects are not easily measurable, and cannot be automatically detected. We strongly believe that ``non-functional'' requirements are fundamental for the design of reusable ontologies and should be always addressed in ontology design methodologies, recommendations and evaluation.




%
%
%
\bibliographystyle{splncs04}
\bibliography{bibliography}

\end{document}